\documentclass[conference]{IEEEtran}
\IEEEoverridecommandlockouts
\usepackage{cite}
\usepackage{amsmath,amssymb,amsfonts}
\usepackage{algorithmic}
\usepackage{graphicx}
\usepackage{textcomp}
\usepackage{xcolor}
\usepackage{booktabs}
\usepackage{color}
\usepackage{url}
\usepackage{bm}
\usepackage{enumitem}
\usepackage{booktabs}
\usepackage[ruled]{algorithm2e}
\def\BibTeX{{\rm B\kern-.05em{\sc i\kern-.025em b}\kern-.08em
    T\kern-.1667em\lower.7ex\hbox{E}\kern-.125emX}}
\begin{document}

\makeatletter
\g@addto@macro\normalsize{%
  \setlength\abovedisplayskip{0.5pt}
  \setlength\belowdisplayskip{0.5pt}
  \setlength\abovedisplayshortskip{0.5pt}
  \setlength\belowdisplayshortskip{0.5pt}
}
\makeatother

\def\BibTeX{{\rm B\kern-.05em{\sc i\kern-.025em b}\kern-.08em
    T\kern-.1667em\lower.7ex\hbox{E}\kern-.125emX}}


\title{Scalable CP Decomposition for Tensor Learning using GPU Tensor Cores}

\author{Zeliang Zhang$^{1}$ , Zhuo Liu$^{1}$ , Susan Liang$^{1}$, Zhiyuan Wang$^{2}$, Yifan Zhu$^{1}$, Chen Ding$^{1}$, Chenliang Xu$^{1}$\\
$^{1}$ School of Computer Science, University of Rochester\\
$^{2}$ School of Computer Science, Huazhong University of Science and Technology}
\maketitle

\begin{abstract}
CP  decomposition is a powerful tool for data science, especially gene analysis, deep learning, and quantum computation.  However, the application of tensor decomposition is largely hindered by the exponential increment of the computational complexity and storage consumption with the size of tensors. While the data in our real world is usually presented as trillion- or even exascale-scale tensors, existing work can only support billion-scale scale tensors. In our work, we propose the Exascale-Tensor to mitigate the significant gap. Specifically, we propose a compression-based tensor decomposition framework, namely the exascale-tensor, to support exascale tensor decomposition. Then, we carefully analyze the inherent parallelism and propose a bag of strategies to improve computational efficiency.   Last, we conduct experiments to decompose tensors ranging from million-scale to trillion-scale for evaluation.  Compared to the baselines, the exascale-tensor supports $8,000$ larger tensors and a speedup up to $6.95\times$.   We also apply our method to two real-world applications, including gene analysis and tensor layer neural networks, of which the numeric results demonstrate the scalability and effectiveness of our method. 
\end{abstract}


\section{Introduction}

Real-world big data are naturally modeled as a multidimensional array, called a tensor. Examples include a time-evolving social network, knowledge base, and gene data. Tensor decomposition is the basis of tensor learning, including graphic analysis, image classification, data mining, etc. Among various tensor decomposition algorithms, Canonical Polyadic (CP) decomposition is mostly used in various real-world applications.  In recent years, the dimensionality of tensors has witnessed a remarkable surge, reaching the scale of trillions ($10^{12}$) to exascale (a billion, $10^{18}$) non-zero elements. Unfortunately, the inherent limitations of primary memory render the conventional CP decomposition algorithm impracticable when dealing with such colossal tensors. 

To confront this formidable challenge, several innovative approaches have been proposed to facilitate large-scale tensor decomposition, among which stand out CDTF \cite{fullyscaleble} and 2PCP \cite{2PCP}, designed specifically for CP tensor decomposition. However, these methods necessitate the input tensors to exhibit an exceptionally sparse structure, accommodating merely millions of non-zero elements. Such stringent sparsity requirements inevitably restrict the scalability of these techniques in real-world applications. Meanwhile, widely employed tensor toolboxes such as Tensor Toolbox \cite{TTB_Software} and TensorLy \cite{tensorly} operate within the constraints of primary memory and thus are ill-suited for handling large, dense tensors. Although the PARACOMP algorithm \cite{paracomp} had shown promise as a compression-based tensor decomposition method, its reported dimensions were limited to a modest $500 \times 500 \times 500$. As the demand for data continues to surge, there arises an urgent need for a scalable CP tensor decomposition approach that can meet the burgeoning requirements of the current big data era.


There are two challenges with large-scale CP tensor decomposition. \underline{First}, when dealing with large-scale tensors, the constraint of limited memory prevents the complete loading of these tensors for computation; \underline{Second}, as the size of the tensors increases, the computational complexity of tensor primitives experiences exponential growth, leading to prohibitively long processing times for tensor decomposition.



To address these challenges, in this paper, we fully explore the potential of compression techniques for tensor decomposition and design a novel method, namely {Exascale-Tensor}, to support the decomposition of large-scale CP tensors up to tensors with exascale elements. First, we trade the storage by the computation with the compression techniques and achieve the large-scale tensor decomposition within limited memory. Second,  we optimize the \textit{Exscale-Tensor} using GPU tensor cores to achieve high-performance computation and support efficient tensor learning applications.    


We summarize the contributions as follows,
\begin{enumerate}
    \item We propose a scalable CP decomposition scheme \textit{Exascale-Tensor}, which trades the computation for storage to achieve large-scale tensor decomposition. 
    \item We propose several strategies to optimize the \textit{Exascale-Tensor} using GPU tensor cores, including efficient memory access, massive parallel compression, and efficient decomposition. 
    \item We conduct extensive experiments to verify the effectiveness of \textit{Exascale-Tensor}. We testify the computational precision and efficiency by ranging the tensor size from $1000 \times 1000 \times 1000$ to $100,000 \times 100,000 \times 100,000$. We also use the optimized \textit{Exascale-Tensor} to support two tensor learning applications, namely gene analysis and tensor neural networks, to demonstrate the scalability of our method. 
\end{enumerate}

The remainder of this paper is organized as follows.
In Section 2, we describe notations and introduce the background on the CP decomposition algorithm. In Section 3, we describe the proposed  \textit{exascale-tensor}. In Section 4, we present the design, implementation, and optimization of the proposed high-performance \textit{exascale-tensor} scheme. In Section 5, we present performance evaluations on the reconstruction error and time consumption. In Section 6, we conclude this paper. 

\section{CP Tensor Decomposition}
\label{tensor_decompositon}


\noindent \textbf{Notations.}
 We use uppercase calligraphic letters to denote third-order tensors, e.g.,  $\mathcal{X} \in \mathbb{R}^{I \times J \times K}$, uppercase boldface letters to denote matrices and lowercase boldface letters to denote vectors, e.g., $\bm{A} \in \mathbb{R}^{I \times J}$ and $\bm{a} \in \mathbb{R}^{I}$. $\bm{X}_{(n)}$ denotes the $n$-mode matricization of a tensor $\mathcal{X}$. We use $\odot$ to denote the Khatri-Rao product and $\otimes$ to denote the tensor (Kronecker) product, which corresponds to the outer product for vectors.

\begin{figure}[t]
    \centering
    \includegraphics[width=\linewidth]{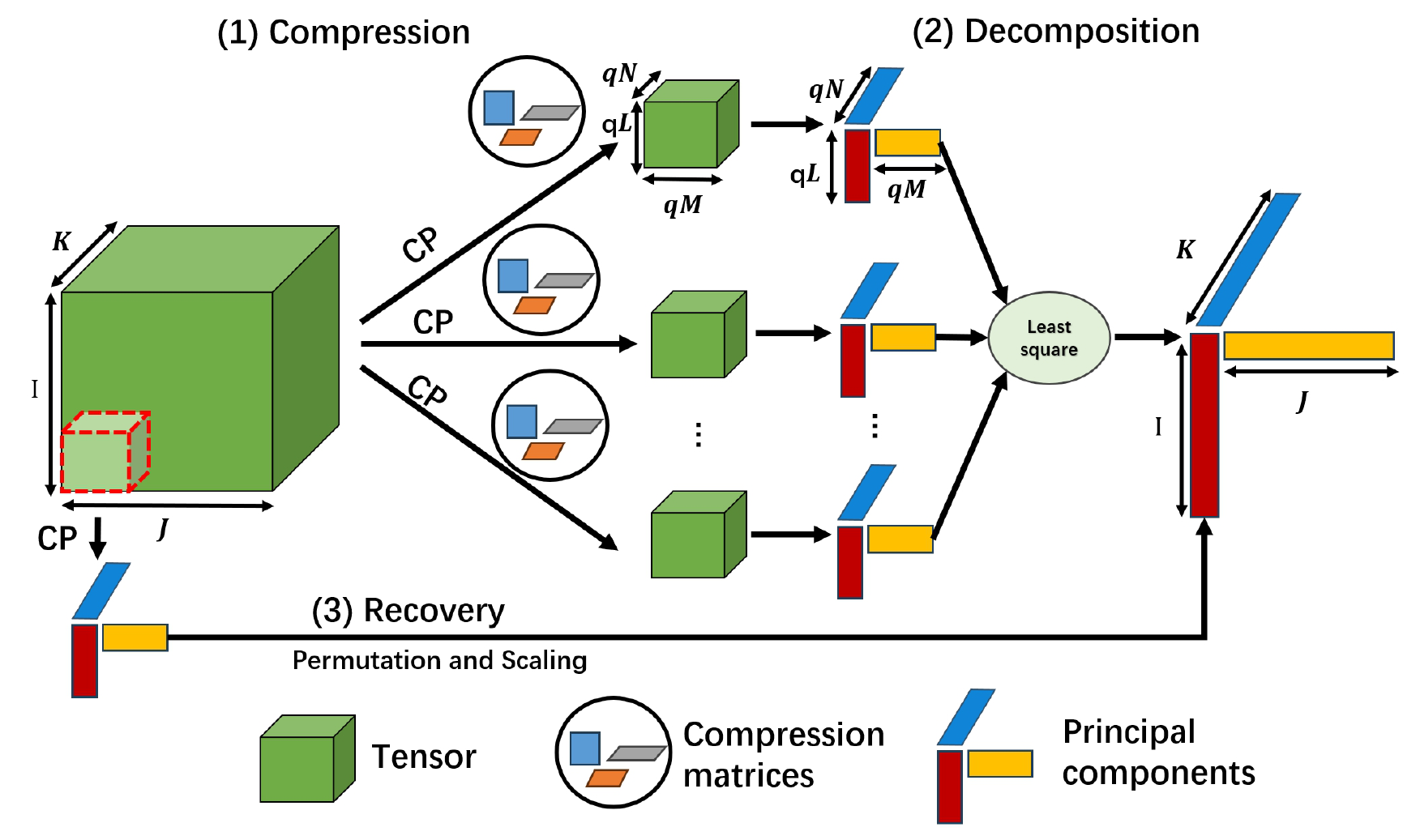}
    \caption{Overview of the proposed \textit{Exascale-Tensor} decomposition. It consists of three stages, namely the compression, decomposition, and recovery stages. }
    \label{fig:overview}
\end{figure}

\noindent \textbf{CP Tensor Decomposition}. For a data tensor $\mathcal{X} \in \mathbb{R}^{I \times J \times K}$, the CP tensor decomposition represents it as a linear combination of $R$ tensor products:
\begin{equation}
\setlength\abovedisplayskip{3pt}
\setlength\belowdisplayskip{3pt}
\label{cp_equation}
    \mathcal{X} = \sum_{r=1}^{R}\bm{a}_{r}\otimes\bm{b}_{r}\otimes\bm{c}_{r}, 
\end{equation}
where $\bm{a}_{r} \in \mathbb{R}^{I}, \bm{b}_{r} \in \mathbb{R}^{J}$, $\bm{c}_{r} \in \mathbb{R}^{K}$ are the principle components derived by the decomposition. The CP tensor decomposition in  (\ref{cp_equation}) can be written in a matrix form,
\begin{equation}
\small
\begin{split}
    &\bm{X_{(1)}}\approx\bm{A}(\bm{C}\odot\bm{B})^{T}, \bm{X_{(2)}}\approx\bm{B}(\bm{C}\odot\bm{A})^{T}, \bm{X_{(3)}}\approx\bm{C}(\bm{B}\odot\bm{A})^{T},
\end{split}
\label{cp_lst_equation}
\end{equation}
where $\bm{A} \in \mathbb{R}^{I \times R}$, $\bm{B} \in \mathbb{R}^{J \times R}$, $\bm{C} \in \mathbb{R}^{K \times R}$ consist of  ${\{\bm{a}_{r}\}}_{r=1}^{R}$, ${\{\bm{b}_{r}\}}_{r=1}^{R}$ and ${\{\bm{c}_{r}\}}_{r=1}^{R}$,  respectively.  


The alternative least squares (ALS) algorithm is widely used to obtain principle component matrices $(\bm{A}, \bm{B}, \bm{C})$, as given in Alg. \ref{alg:baseline CP}. First, ($\bm{A},\bm{B},\bm{C}$) are randomly initialized in line $1$, \textit{e.g.}, following an identical independent normal distribution. Then, the while-loop in line $2$--$6$ alternatively fixes two matrices and solves the least square problems, as shown in Eq. (\ref{cp_lst_equation}), to obtain the third matrix. The while-loop stops until the reconstruction error is below a preset threshold or the maximum number of iterations is reached.  

However, the above updates of $\bm{A}$, $\bm{B}$ and $\bm{C}$ require to load the full tensor $\mathcal{X}$ into the main memory. Thus, the capacity of memory becomes the bottleneck, restricting the supported tensor size. The compression theory in tensor decomposition~\cite{paracomp} provides a promising way to derive the original principle components by decomposing the proxy tensors with reduced size. This motivates us to further utilize this technique to trade the computation with storage for supporting large-scale tensor decomposition.

\section{Exascale CP Tensor Decomposition}
\label{SEC_IMPLEMENTATION}

\begin{algorithm}[t] 
\caption{ALS algorithm for CP decomposition} 
\label{alg:baseline CP} 
\begin{algorithmic}[1] 
\REQUIRE a third-order tensor $\mathcal{X}$ of size $I \times J \times K$, rank $R$, and maximum number of iterations $T$,
\ENSURE 
matrices $(\bm{A}, \bm{B}, \bm{C})$.\\
\STATE Initialize $\bm{A} \in \mathbb{R}^{I \times R}, \bm{B} \in\mathbb{R}^{J \times R}, \bm{C} \in \mathbb{R}^{K \times R}$, $t=0$,
\WHILE{$t \leq T$ and NOT converged}
    \STATE $\bm{A} \gets  \bm{X}_{(1)}(\bm{C}\odot\bm{B})(\bm{C}\bm{C}^{T}*\bm{B}\bm{B}^{T})$,
    \STATE $\bm{B} \gets \bm{X}_{(2)}(\bm{C}\odot\bm{A})(\bm{C}\bm{C}^{T}*\bm{A}\bm{A}^{T})$,
    \STATE $\bm{C} \gets  \bm{X}_{(3)}(\bm{B}\odot\bm{A})(\bm{B}\bm{B}^{T}*\bm{A}\bm{A}^{T})$,
\ENDWHILE
\RETURN $(\bm{A}, \bm{B}, \bm{C})$. 
\end{algorithmic}
\end{algorithm}

As shown in Fig. \ref{fig:overview}, the proposed \textit{Exascale-Tensor} consists of three stages, namely the compression, decomposition, and recovery stages.  First, we randomly compress the large tensor into multiple small tensors. Then, we use conventional CP decomposition to decompose the small tensors into principal components and derive the approximated principal components of the original large tensor from the principal components collected belonging to the proxy tensor. Last, we recover the original principal components of the original large-scale tensor. 

\noindent \textbf{Compression}. In line 1--2 of  Alg. \ref{alg:Twice compression Framwork}, we compress the large tensor into multiple proxy  tensors with reduced size. Generally, to compress tensor $\mathcal{X} \in \mathbb{R}^{I \times J \times K}$ into $\mathcal{Y} \in \mathbb{R}^{L \times M \times N}$, the $(i,j,k)$-th element of $\mathcal{Y}$ is computed as follows: 
\begin{equation}
\small
\setlength\abovedisplayskip{3pt}
\setlength\belowdisplayskip{3pt}
\label{compression_equation}
    \mathcal{Y}_{lmn}=\sum\limits_{i=1}^{I}\sum\limits_{j=1}^{J}\sum\limits_{k=1}^{K}\bm{U}_{li}\bm{V}_{mj}\bm{W}_{nk}\mathcal{X}_{ijk},
\end{equation}
where $\bm{U}\in\mathbb{R}^{L \times I} , \bm{V}\in\mathbb{R}^{M \times J}, \bm{W}\in\mathbb{R}^{N \times K}$ are random matrices sampled from the normal distribution. We denote (\ref{compression_equation}) as $\mathcal{Y}=\text{Comp}(\mathcal{X}, \bm{U}, \bm{V}, \bm{W})$ for simplicity. 

At the compression stage, we first generate $P$ random matrices from the normal distribution, $\{\bm{U}_{p},\bm{V}_{p},\bm{W}_{p}\}_{p=0}^{P}$. Among them, $\bm{U}_{p}\in \mathbb{R}^{L \times I},\bm{V}_{p}\in \mathbb{R}^{M \times J},\bm{W}_{p}\in \mathbb{R}^{N \times K}$, where we denote $L\times M \times N$ as the reduced size of tensors We compute the proxy tensors $\mathcal{Y}_{p}=\text{Comp}(\mathcal{X}, \bm{U}_{p}, \bm{V}_{p}, \bm{W}_{p})$, $p=1,2,...,P$. It implicitly involves two levels of compression, reducing the size of tensors from $I\times J \times K$ to $L \times M \times N$.

\begin{algorithm}[t] 
\caption{Exascale-Tensor Scheme.} 
\label{alg:Twice compression Framwork} 
\begin{algorithmic}[1] 
\REQUIRE ~~\\ 
tensor $\mathcal{X} \in \mathbb{R}^{I \times J \times K}$ with rank F; reduced sizes of the tensor, $L, M, N$;  the number of proxy tensors $P$; the number of common columns $S$.
\ENSURE 
principal component matrices $(\bm{A}, \bm{B}, \bm{C})$.\\
\STATE Randomly generate matrices $\bm{U}_{p} \in \mathbb{R}^{L \times I},\bm{V}_{p} \in \mathbb{R}^{M \times J},\bm{W}_{p} \in \mathbb{R}^{N \times K}$, $p=1,2, .., P$, and set the first $S$ columns the same,

\STATE $\mathcal{Y}_{p} \gets$ compress $\mathcal{X}$ using $(\bm{U}_{p},\bm{V}_{p},\bm{W}_{p})$, for $p = 1, .., P$,
\label{code:fram:factorization}
\FOR{$p=1$ to $P$} 
\STATE $(\bm{A}_{p}, \bm{B}_{p}, \bm{C}_{p})$ $\gets$ rank-R CP decomposition  of $\mathcal{Y}_{p}$ via Alg. \ref{alg:baseline CP},
\STATE $(\bm{A}_{p},\bm{B}_{p},\bm{C}_{p})\gets$  divide each column of $(\bm{A}_{p},\bm{B}_{p},\bm{C}_{p})$ by the maximum of the first $S$ rows, respectively,
\STATE $\bm{\Pi}_{p}$ $\gets$  $\arg\max_{\bm{\Pi}} \text{Tr}(\bm{A}_{1}(1:S, :)^{T}\bm{A}_{p}(1:S, :)\bm{\Pi})$,
\STATE $(\bm{A}_{p},\bm{B}_{p},\bm{C}_{p}) \gets$  $(\bm{A}_{p}\bm{\Pi}, \bm{B}_{p}\bm{\Pi}, \bm{C}_{p}\bm{\Pi})$,
\ENDFOR
\label{code:fram:join}
\STATE $(\bm{A}\bm{\Pi}\bm{\Sigma}, \bm{B}\bm{\Pi}\bm{\Sigma}, \bm{C}\bm{\Pi}\bm{\Sigma})$ $\gets$ solve  the least squares problem (\ref{lsq}) per mode,
\STATE $(\hat{\bm{A}},\hat{\bm{B}},\hat{\bm{C}})$ $\gets$ rank-R CP decomposition of the sampled tensor $\mathcal{B} \in \mathbb{R}^{b \times b \times b}$ in $\mathcal{X}$,
\STATE $\bm{\Pi},\bm{\Sigma}$ $\gets$ match $\hat{\bm{A}}$ with the first $d$ rows of $\bm{A}\bm{\Pi}\bm{\Sigma}$ using the Hungarian algorithm.
\STATE $\bm{A}$ $\gets$ $(\bm{A}\bm{\Pi}\bm{\Sigma})(\bm{\Pi}\bm{\Sigma})^{\dagger}$,
\STATE Repeat line 11-12 for $\bm{B}$ and $\bm{C}$, respectively,
\label{code:fram:join}

\RETURN $(\bm{A}, \bm{B}, \bm{C})$.
\end{algorithmic}
\end{algorithm}
\label{sec:decomposition}
\noindent \textbf{Decomposition}. The decomposition stage is involved in line 3--9 of Alg. \ref{alg:Twice compression Framwork}. On the decomposition stage, each proxy tensor $\mathcal{Y}_{p}$ will be independently decomposed to $(\bm{A}_{p},\bm{B}_{p},\bm{C}_{p})$, $p=1,2,..,P$, using the conventional CP decomposition shown in Alg.\ref{alg:baseline CP}. Due to the property of the Kronecker product \cite{kr_property}, we have the equation $\bm{A}_{p}=\bm{U}_{p}\bm{A}\bm{\Pi}_{p}\bm{\Sigma}_{p}$,  $p=1,2,...,P$.  Here, we apply the Hungarian algorithm \cite{HA} to the trace maximization problem to remove the column permutation matrix $\bm{\Pi}_{p}$ to $\bm{\Pi}$  and divide each column by its maximum of the first $S$ rows to get rid of the magnitude scaling matrix $\bm{\Sigma}_{p}$ to $\bm{\Sigma}$, where we preliminarily set $S$ shared columns in the random matrices $\bm{U}_{p}, p=1,2,...,P$, at the compression stage. Finally, solve a  least squares problem,
\begin{equation}
\label{lsq}
\left[
\begin{array}{c}
     \bm{A}_{1}\\
     \vdots    \\
     \bm{A}_{P}\\
\end{array}
\right]
=
\left[
\begin{array}{c}
    \bm{U}_{1} \\ 
    \vdots    \\
    \bm{U}_{P}
\end{array}
\right]
\bm{A}\bm{\Pi}\bm{\Sigma}
\end{equation}
to obtain $\bm{A}\bm{\Pi}\bm{\Sigma}$,  and similarly for $\bm{B}\bm{\Pi}\bm{\Sigma}$ and $\bm{C}\bm{\Pi}\bm{\Sigma}$. 

\noindent \textbf{Recovery}. The recovery stage is involved in line 10--13 of Alg. \ref{alg:Twice compression Framwork}. Due to the uniqueness of CP decomposition, we propose to sample a small tensor from the original large-scale tensor and use the principal components of this small tensor to derive $\bm{\Pi}$ and $\bm{\Sigma}$.  Suppose the sampled tensor  $\mathcal{B} \in \mathbb{R}^{b \times b \times b}$, it has the decomposed principal components  $(\hat{\bm{A}},\hat{\bm{B}},\hat{\bm{C}})$, which ara different with the first $b$ rows of $\bm{B}\bm{\Pi}\bm{\Sigma}$ and $\bm{C}\bm{\Pi}\bm{\Sigma}$ in permutation and scaling. Thus, we employ the Hungarian algorithm again to match the columns between these two sets of principle components and get rid of the $\bm{\Pi}\bm{\Sigma}$.


\section{High-performance Exascale Tensor}

\subsection{Efficient Memory Access}
\label{sec:mem}
In \textit{Exascale-Tensor}, we need to frequently access tensors along different dimensions, \textit{, that is,}, $\bm{X}_{(1)},\bm{X}_{(2)},\bm{X}_{(3)}$.  We propose to store the data in the column-major format to avoid explicit conversions between tensors and matrices in GPU memory. $\bm{X}_{(1)}$ in column-major  can be directly fetched from the storage of $\mathcal{X}$.  $\bm{X}_{(2)}$ and $\bm{X}_{(3)}$ in row-major can be accessed by transposing each slice of $\mathcal{X}$ and vectorizing $\mathcal{X}$, respectively.   Furthermore, the column-major storage has good support for GPU functions in NVIDIA CUDA libraries, which is suitable for the high-performance implementation of \textit{ Exascale-Tensor}.

\subsection{Mixed-precision Acceleration using GPU Tensor Cores}
There are many tensor learning primitives in the \textit{Exascale-Tensor}, including the Khatri-Rao product, tensor product, matrix multiple multiplications, \textit{etc}. With the increasing size of tensors, the computational complexity of these operations grows exponentially, which hinders the applications. Fortunately, the tensor learning primitives have a high degree of parallelism, which is amenable for GPU computations. Some studies also point out that the tensor learning primitives can be transformed into matrix computation and mapped onto GPU tensor cores to achieve significant improvement in efficiency. Thus, we use GPU tensor cores to support the tensor computation involved, thus accelerating the \textit{Exascale-Tensor}.

However, GPU tensor cores provide acceleration by using mixed-precision ($\text{FP}_{16}\times \text{FP}_{16} + \text{FP}_{32} \rightarrow \text{FP}_{32}$) for $4 \times 4$ matrix multiplication, which could cause a high cumulative rounding error in \textit{Exascale-Tensor}, especially at the compression stage. To mitigate the precision loss, we propose to factorize FP32 format data to FP16 format data and its residual error from conversion. We factorize the  compression $\mathcal{Y}{p}=\text{Comp}(\mathcal{X}, \bm{U}_{p},\bm{V}_{p},\bm{W}_{p})$ can be factorize as follows, 
\begin{equation}
    \label{residual error compression}
    \small
\setlength\abovedisplayskip{3pt}
\setlength\belowdisplayskip{3pt}
    \begin{aligned}
    \mathcal{Y}_{p}
    \approx &\text{Comp}(\mathcal{X}^{16}, \bm{U}^{16}_{p},\bm{V}^{16}_{p},\bm{W}^{16}_{p})+\text{Comp}(\mathcal{X}^{16}, \widetilde{\bm{U}^{16}_{p}},\bm{V}^{16}_{p},\bm{W}^{16}_{p})\\
    &\text{Comp}(\mathcal{X}^{16}, \bm{U}^{16}_{p},\widetilde{\bm{V}^{16}_{p}},\bm{W}^{16}_{p})+\text{Comp}(\mathcal{X}^{16}, \bm{U}^{16}_{p},{\bm{V}^{16}_{p}},\widetilde{\bm{W}^{16}_{p}})\\
    &\text{Comp}(\widetilde{\mathcal{X}^{16}},\bm{U}^{16}_{p},\bm{V}^{16}_{p},\bm{W}^{16}_{p}),
\end{aligned}
\end{equation}
where $\bm{U}^{16}_{p}$, $\bm{V}^{16}_{p}$, $\bm{W}^{16}_{p}$, $\mathcal{X}^{16}$ are stored in FP16 format, $\widetilde{\bm{U}^{16}_{p}}$, $\widetilde{\bm{V}^{16}_{p}}$, $\widetilde{\bm{W}^{16}_{p}}$, $\widetilde{\bm{W}^{16}_{p}}$ are the residual error for the precision conversion from FP32 to FP16 and we only consider the computations with first-order residual error and ignore the computations for the high-order residual error. 


\subsection{Massive Parallel Compression}

 \begin{figure}
     \centering
     \includegraphics[width=\linewidth]{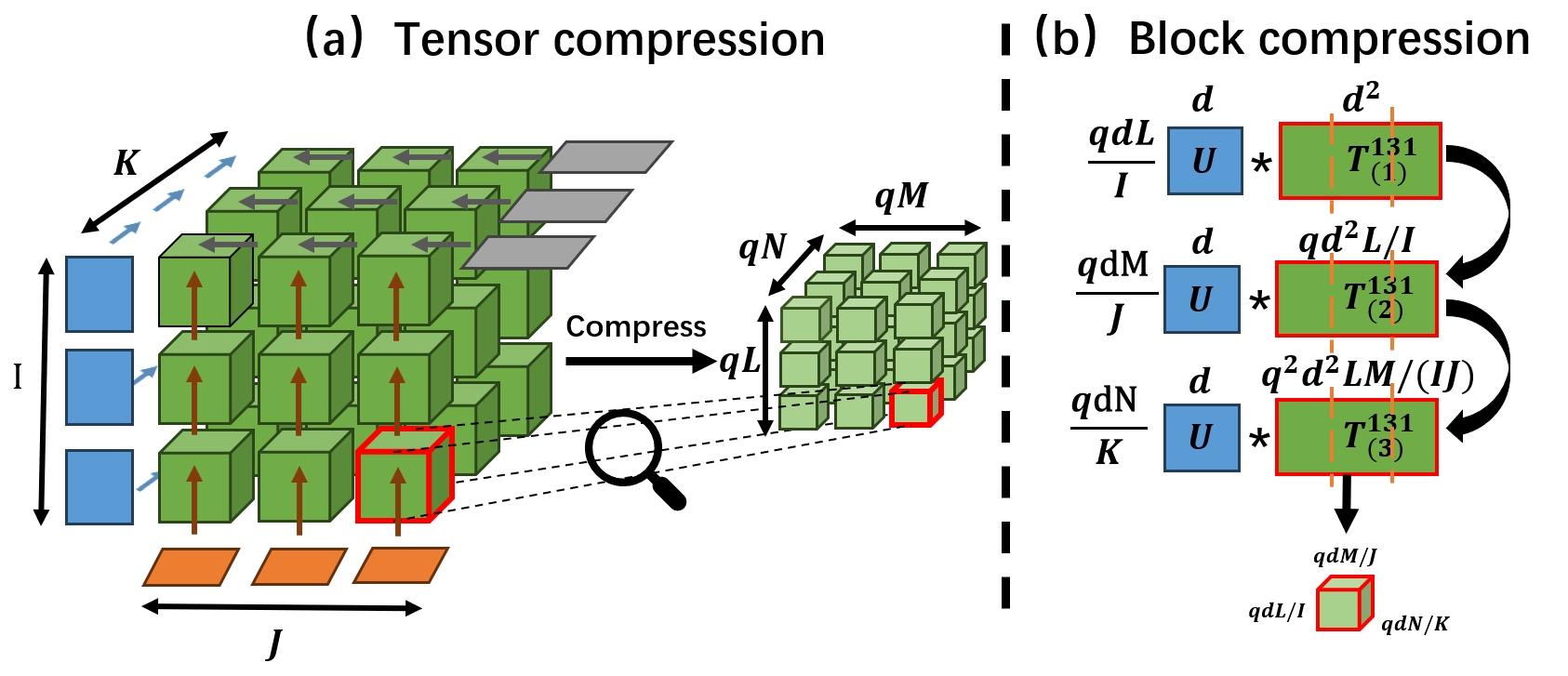}
     \caption{Illustration of the tensor compression in block computation. In (a), we split the original tensor into multiple blocks, compress each tensor block, and merge the results into the final compressed tensor. In (b), we employ the compression matrices to compress each dimension of the tensor block, where the efficient memory access technique is utilized to avoid the explicit conversions between tensor and matrices. We also split the matricized tensor into multiple blocks and map them onto GPU tensor cores for batch computation. }
     \label{fig:tensor compression}
 \end{figure}
 
 Benefiting from the inherent large parallelism in tensor operations, we compress the large-scale tensor by computing it in blocks, as shown in Fig. \ref{fig:tensor compression}. Supposing $I,J$ are multiplies of $d$, for the large-scale tensor $\mathcal{X}$ with size $I \times J \times K$,  we split it into different tensor slices $\mathcal{T}^{ijk} \in \mathbb{R}^{d \times d \times d}$,  $i \in [I], j \in [J], k \in [K]$.  The elements of  $\mathcal{T}^{ijk}$ are composed of $\mathcal{X}(id:(i+1)d,jd:(j+1)d,kd:(k+1)d)$, which are amenable for $d^2$ GPU threads for continuous column-major memory access.

To compress $\mathcal{T}^{iik}$,  we fetch the compression block matrices $\bm{U}(:,id:(i+1)d), \bm{V}(:,jd:(j+1)d), \bm{W}(:,kd:(k+1)d)$, and multiply the matricized tensor $\bm{T}^{ijk}_{(1)},\bm{T}^{ijk}_{(2)},\bm{T}^{ijk}_{(3)}$ to compress each dimension, \textit{i.e.}, $I,J,K$, respectively. It should be noted that the explicit conversion between $\mathcal{T}^{iik}$ and $\bm{T}^{ijk}_{(1)},\bm{T}^{ijk}_{(2)},\bm{T}^{ijk}_{(3)}$ can be avoided with the introduced technique in Section \ref{sec:mem}. Here, we omit the proxy tensor index $p$ for simplicity. 

There is massive parallelism at the compression stage. On the one hand, note that the compressions of all tensor blocks are independent, which is amenable for parallel computation. On the other hand, in terms of each block computation, it consists of three times the multiplication of the fat matrix due to $\mathcal{T}$. To better use the GPU tensor cores, in each multiplication of matrices, we first divide the matricized $\mathcal{T}$ into multiple blocks, of which the dimensions are multiplied by $8$. Then, we employ the GPU tensor cores to compute the fat matrix multiplications in batch. Last, we contact the results of batch matrix multiplication as the resulting matricized tensor.

\subsection{Efficient Decomposition}
The least square solution of $\bm{A}\bm{\Pi}\bm{\Sigma}$ in (\ref{lsq}) requires the memory footprint of  $\mathcal{O}\left(I \times \left(1 +  PL\right)\right)$. On the one hand, as identified in \cite{paracomp}, to meet the precision requirement with the compression technique, $P\ge \frac{I-2}{L-2}$, which indicates that a larger compression ratio ($L\downarrow$) needs a larger number of proxy tensors for principal component estimation ($P\uparrow$). Besides, a larger compression ratio brings a larger precision loss. Thus, it poses a great challenge to reduce the memory footprint to improve the decomposition efficiency as well as having an acceptable precision loss. 

We propose a constructive approach to address this challenge. We integrate the compressed sensing technique~\cite{donoho2006compressed} to \textit{Exascale-Tensor}. Specifically, at the compression stage, we generate the compression matrices by $\bm{U}_{p}=\bm{U'}_{p}\bm{U}$, $\bm{V}_{p}=\bm{V'}_{p}\bm{V}$, and $\bm{W}_{p}=\bm{W'}_{p}\bm{W}$, where we have $\bm{U'}_{p} \in \mathbb{R}^{L \times \alpha L}$, $\bm{V'}_{p} \in \mathbb{R}^{M \times \beta M}$, $\bm{W'}_{p} \in \mathbb{R}^{N \times \gamma N}$, $\bm{U} \in \mathbb{R}^{\alpha L \times I}$, $\bm{V} \in \mathbb{R}^{\beta M \times J}$, $\bm{W} \in \mathbb{R}^{\gamma N \times K}$, and $\alpha,\beta,\gamma>1$. With this construction process, the compression stage is implicitly transformed into a twice-stage compression, namely from $I\times J 
\times K$ to $\alpha L \times \beta M \times \gamma N$, then to $L \times M \times N$. Correspondingly, the original decomposition stage is transformed into a two-stage factorization process, where we first derive the $\bm{U}\bm{A}\bm{\Pi}\bm{\Sigma}$, then obtain the $\bm{A}\bm{\Pi}\bm{\Sigma}$.

Such a design has two advantages. First, the integration of the compressed sensing technique allows us to use sparse matrices $(\bm{U},\bm{V},\bm{W})$, which provides more efficiency to the compression stage. Second, the implicit first-time compression allows a larger compression ratio $(L\downarrow)$ with remaining the number of proxy tensors $P$, leading to a reduced memory footprint. Besides, due to the sparsity of matrices $(\bm{U},\bm{V},\bm{W})$, the introduced additional factorization stage, from $\bm{U}\bm{A}\bm{\Pi}\bm{\Sigma}$ to  $\bm{A}\bm{\Pi}\bm{\Sigma}$, can be efficiently solved by the $L_{1}$ constraint minimization algorithm~\cite{dadkhah2010compressive}, which is faster and more numerical stable than least square method.


\section{Performance Evaluations}


In this section, we will provide numerical experiment results to show the performance of our work on various scales of tensors from one hundred million scale to exascale.

\subsection{Experimental Settings}

We conducted our experiments on a server that has two Intel(R) Xeon(R) Gold 5118 CPUs. Each of the CPUs has 12 cores @2.30GHz supporting 24 hardware threads. There is a Titan RTX GPU, which consists of 24 GB of device memory. There are 256 GB DDR4 memories on the server. We show the mean squared error (MSE) to measure the accuracy and speedups of baseline versus optimization schemes. The speedups are defined as  $\frac{\text{baseline running time}}{\text{optimized running time}}$.

\textbf{We will first show the results of our algorithm on decomposing various tensors with ranging sizes}. 

For the evaluation of factorization for dense tensors, we test cases on scales ranging from one hundred million to one trillion, which approaches the limitations restricted by the device. For each case, we first generate the mode matrices $\mathit{A} \in \mathbb{R}^{I \times F}$,$\mathit{B} \in \mathbb{R}^{J \times F},\mathit{C} \in \mathbb{R}^{K \times F}$ from an independent normal distribution to generate the huge tensor $\mathcal{X} \in \mathbb{R}^{I \times J \times K}$ where we set $I=J=K$ ranging from 1000 to 10000 and the rank F is set to be a constant small value 5. On the compression stage, we set the size of compressed tensor cubes $\mathcal{Y}_{p} \in \mathbb{R}^{L \times M \times N}$ as the same value $L=M=N=50$. The size of one tensor block $\mathit{B} \in \mathbb{R}^{d_{1} \times d_{2} \times d_{3}}$ on the compression stage is set as $d_{1}=d_{2}=d_{3}=500$ while it is set as $d_{1}=20,d_{2}=40$ and $d_{3}=60$ on the scaling and permutation stage.  We set P as $\max(\frac{I-2}{L-2},\frac{J}{M},\frac{K}{N})$+10 to avoid the situation that if the CP tensor decomposition of one or a little more compression replicas can't converge on the factorization stage,  drop it (them) in time. 

For the evaluation of factorization for sparse tensors, we test cases on scales ranging from one to six hundred million. The generation of large-scale tensors $\mathcal{X} \in \mathbb{R}^{I \times J \times K}$ is similar to that for dense tensors but sets the number of non-zero elements in each mode matrix as one hundred of the dimension. We denote the  size of compression replica $\mathcal{Y} \in \mathbb{R}^{L \times M \times N}$ as $L=\frac{I}{10},M=\frac{J}{10},N=\frac{K}{10}$. 

For the evaluation of exascale tensors, we test the case for exascale tensors by ranging the sparsity. For each case, we range the number of non-zero elements of the mode matrices to control the sparsity of the exascale tensors. 

\textbf{Then, we will use our proposed algorithm to support two tensor learning applications, namely the CP tensor layer for neural networks~\cite{lebedev2015speeding} and gene analysis using CP decomposition~\cite{hore2016tensor}.}

For the CP tensor layer, we use respectively our algorithm and the CP decomposition provided by Matlab~\cite{Toolbox2018} and TensorLy~\cite{kossaifi2016tensorly} to decompose the ResNet-34~\cite{he2016deep} and fine-tune it on the CIFAR-10 dataset. For evaluation, we report the performance degradation at the factorization stage and the classification accuracy after the fine-tuning. 

For gene analysis, following the same setting in the previous study~\cite{liu2022high}, we use our algorithm to decompose the gene data. We report the loss of the information by the decomposition and the used time for factorization. 
\begin{figure}[h]
    \centering
    \includegraphics[width=0.8\linewidth]{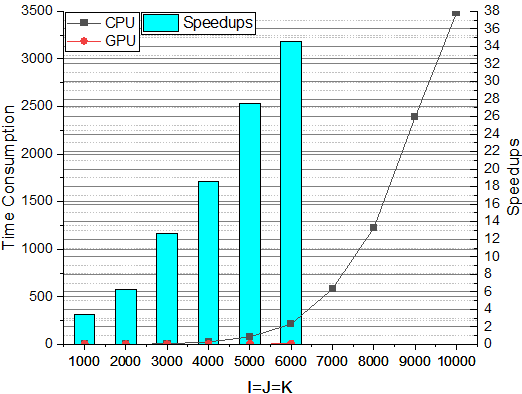}
    \caption{Comparison of time performance between baseline (CPU) and optimized version using GPU tensor cores (GPU).}
    \label{sparsetime}
\end{figure}
\subsection{Results on Tensor Decomposition}

\begin{figure}[h]
    \centering
    \includegraphics[width=0.8\linewidth]{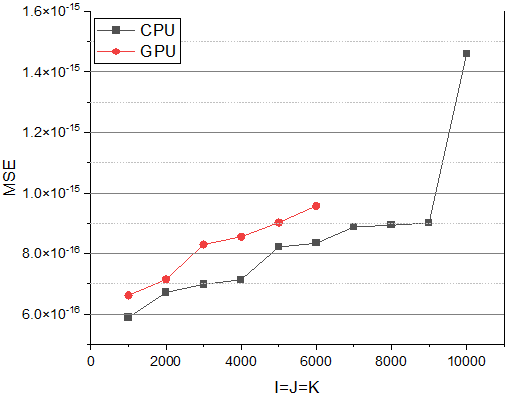}
    \caption{Comparison of reconstruction mean squared error (MSE) between baseline (CPU) and optimized version using GPU tensor cores (GPU).}
    \label{sparseerror}
\end{figure}

\noindent \textbf{Sparse Tensor Decomposition}.  Figure \ref{sparsetime} shows the comparison of time performance between the baseline and the optimized versions. Compared to the baseline,   the optimized version on GPUs achieves an average of $17.17 \times$ speedups with up to $34.60 \times$ speedups. For tensor of size $6,000 \times 6,000 \times 6,000$, we reduce the baseline time of three and a half hours to nearly 6 minutes.

Figure \ref{sparseerror} shows the comparison of MSE between the baseline and the optimized version. It can be seen that all the MSEs are increasing as the scale increases,  but all of them control the error under the magnitude of $10^{-15}$.  

Using only one compression replica and compressive sensing method makes the decomposition of sparse tensors heavily restricted by the factorization stage for the limited VRAM. In our experiment, the max size of tensor that can handle on GPU is $6,000 \times 6000 \times 6000$ when the compression rate is denoted as $10$ on the compression stage.

\begin{figure}[t]
    \centering
    \includegraphics[width=0.8\linewidth]{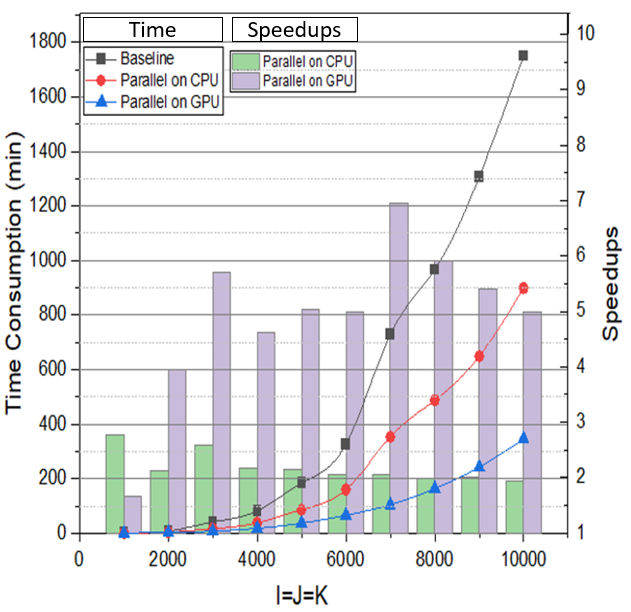}
    \caption{Comparison of time performance between baseline (Baseline) and optimized versions with parallel techniques using MPI (Parallel on CPU) and GPU tensor cores (Parallel on GPU).}
    \label{densetime}
\end{figure}

\begin{figure}[t]
    \centering
    \includegraphics[width=0.8\linewidth]{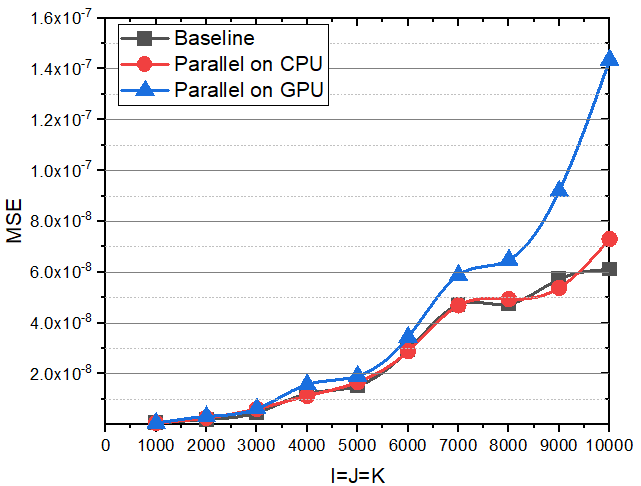}
    \caption{Comparison of reconstruction mean squared error(MSE) between baseline (Baseline) and optimized versions with parallel techniques using MPI (Parallel on CPU) and GPU tensor cores (Parallel on GPU).}
    \label{denseerror}
\end{figure}

\noindent \textbf{Dense Tensor Decomposition}. Figure \ref{densetime} shows the comparison of time performance between the baseline and the optimized version. Compared to the baseline,  the optimized version on CPUs achieves an average of $2.18 \times$ speedups with up to $2.77 \times$ speedups, and the optimized version on GPUs achieves an average of $4.92 \times$ speedups with up to $6.95 \times$ speedups. For a trillion-scale tensor of size $10,000 \times 10,000 \times 10,000$, we reduce the baseline time of nearly half a day to no more than 6 hours.

Figure \ref{denseerror} shows the comparison of MSE between the baseline and the optimized versions. It can be seen that all the MSEs are increasing as the scale increases,  but all of them control the error under the magnitude of $10^{-7}$. Compared to the baseline, the parallel scheme on CPUs is almost the same with more than $2 \times$ speedups. For the parallel scheme on GPUs, it trades just a little more loss for the significant acceleration.

\begin{figure}[t]
    \centering
    \includegraphics[width=0.8\linewidth]{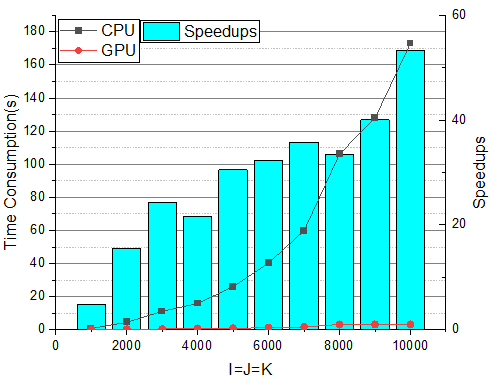}
    \caption{ time consumption between baseline (Baseline) and optimized versions with parallel techniques using MPI (Parallel on CPU) and GPU tensor cores (Parallel on GPU).}
    \label{twostagetime}
\end{figure}

\begin{figure}[t]
    \centering
    \includegraphics[width=0.8\linewidth]{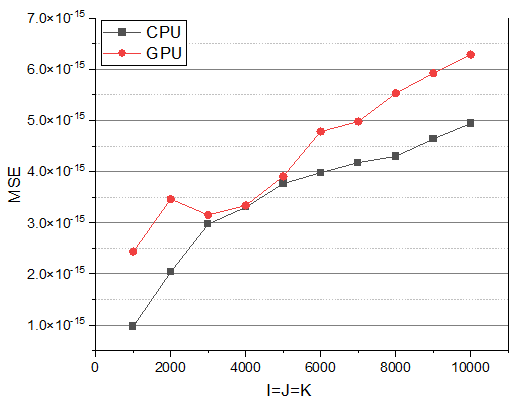}
    \caption{Comparison of reconstruction mean squared error(MSE) between baseline (Baseline) and optimized versions GPU tensor cores (GPU).}
    \label{twostageerror}
\end{figure}

\noindent \textbf{Exascale-Tensor Decomposition}. Figure \ref{twostagetime} shows the comparison of time performance between the baseline and the optimized version. Compared to the baseline,  the optimized version on GPUs achieves an average of $56.52 \times$ speedups with up to $172.98 \times$ speedups. For a trillion-scale tensor of size $10,000 \times 10,000 \times 10,000$, we reduce the baseline time of nearly half $18$ hours to merely $20$ minutes.

Figure \ref{twostageerror} shows the comparison of MSE between the baseline and the optimized version. It can be seen that all the MSEs are increasing as the scale increases,  but all of them control the error under the magnitude of $10^{-14}$. 

Compared with directly using the compressive sensing method with only one replica or factorization of large-scale tensors with no assumption of sparsity, \textit{Exascale-Tensor} can factorize large-scale tensors with more efficiency and less loss for precision. 

\subsection{Tensor Learning Applications}

\begin{table}[t]
\caption{Classification accuracy and time consumption of the CP decomposition on ResNet-34 trained on   CIFAR-10.}\label{tab: cp_nn}
\centering
\begin{tabular}{c|ccc}
\toprule
Methods       & Matlab & TensorLy & \textbf{Ours} \\
\hline
Accuracy ($\%$) &          63.7   &    59.2      &     \textbf{67.8}               \\
Time (seconds)          &         133      &   125       &     \textbf{91}                 \\
\hline
\end{tabular}%
\end{table}

\noindent \textbf{CP tensor layer for neural networks}.  The CP decomposition can be applied to compress the neural networks with acceptable performance degradation. We apply our proposed method to compress the ResNet-34 trained on the CIFAR-100 dataset. We take the Tensor Toolbox in Matlab and TensorLy as the baseline methods for comparison.  As shown in \ref{tab: cp_nn}, after the factorization, all the ResNet demonstrates a performance degradation. While the runner-up method provided by Matlab achieves an improvement of $4.1\%$, our method has the best classification accuracy of $67.8\%$, indicating less information lost at the factorization stage. 

\noindent \textbf{Gene analysis using CP decomposition}. The gene data is usually modeled as a third-order 'individual-tissue-gene' tensor.  The factorization of gene data using CP tensor decomposition can be used to analyze gene expression across multiple tissues and uncover biological relationships. However, the large volume of gene data and computational complexity limit this method's application due to high computational demands and memory constraints. While previous work fails to decompose the large-scale gene database, our proposed method can be utilized to factorize it into multiple principal components for analysis. The relative error after the factorization is $1.4\%$, and the running time is $137$ seconds, which shows great potential for big data analysis.

\section{Conclusions}

In this paper, we propose a \textit{Exascale-Tensor} scheme for CP tensor decomposition that can support exascale-scale tensors. We provide a bag of optimization techniques to accelerate the scheme, including efficient memory access, mixed-precision acceleration, and efficient decomposition. In our experiment, we test various tensor sizes ranging from million scale to exascale and obtain a low mean squared error. Additional experiments on two tensor learning applications, including the CP tensor layer for neural networks and gene analysis, demonstrate the scalability of our methods. 

\bibliographystyle{IEEEbib}
\bibliography{mybibfile}

\begin{thebibliography}{10}

\bibitem{fullyscaleble}
Kijung Shin, Lee Sael, and U~Kang,
\newblock ``Fully scalable methods for distributed tensor factorization,''
\newblock {\em IEEE Transactions on Knowledge and Data Engineering}, vol. 29, no. 1, pp. 100--113, 2016.

\bibitem{2PCP}
Xinsheng Li, Shengyu Huang, K~Sel{\c{c}}uk Candan, and Maria~Luisa Sapino,
\newblock ``2pcp: Two-phase cp decomposition for billion-scale dense tensors,''
\newblock in {\em 2016 IEEE 32nd International Conference on Data Engineering (ICDE)}. IEEE, 2016, pp. 835--846.

\bibitem{TTB_Software}
Brett~W. Bader, Tamara~G. Kolda, et~al.,
\newblock ``Matlab tensor toolbox version 3.1,'' Available online, June 2019.

\bibitem{tensorly}
Jean Kossaifi, Yannis Panagakis, Anima Anandkumar, and Maja Pantic,
\newblock ``{TensorLy}: Tensor learning in python,''
\newblock {\em Journal of Machine Learning Research}, 2019.

\bibitem{paracomp}
Nicholas~D Sidiropoulos, Evangelos~E Papalexakis, and Christos Faloutsos,
\newblock ``Parallel randomly compressed cubes: A scalable distributed architecture for big tensor decomposition,''
\newblock {\em IEEE Signal Processing Magazine}, vol. 31, no. 5, pp. 57--70, 2014.

\bibitem{kr_property}
J.~Brewer,
\newblock ``Kronecker products and matrix calculus in system theory,''
\newblock {\em IEEE Trans. Circuits Syst.}, vol. 19, pp. 772--781, 1978.

\bibitem{HA}
Harold~W. Kuhn,
\newblock ``The hungarian method for the assignment problem,''
\newblock {\em Naval Research Logistics Quarterly}, pp. 83--97, 1955.

\bibitem{donoho2006compressed}
David~L Donoho,
\newblock ``Compressed sensing,''
\newblock {\em IEEE Transactions on information theory}, vol. 52, no. 4, pp. 1289--1306, 2006.

\bibitem{dadkhah2010compressive}
MR~Dadkhah, Shahram Shirani, and M~Jamal Deen,
\newblock ``Compressive sensing with modified total variation minimization algorithm,''
\newblock in {\em 2010 IEEE International Conference on Acoustics, Speech and Signal Processing}. IEEE, 2010, pp. 1310--1313.

\bibitem{lebedev2015speeding}
V~Lebedev, Y~Ganin, M~Rakhuba, I~Oseledets, and V~Lempitsky,
\newblock ``Speeding-up convolutional neural networks using fine-tuned cp-decomposition,''
\newblock in {\em 3rd International Conference on Learning Representations, ICLR 2015-Conference Track Proceedings}, 2015.

\bibitem{hore2016tensor}
Victoria Hore, Ana Vinuela, Alfonso Buil, Julian Knight, Mark~I McCarthy, Kerrin Small, and Jonathan Marchini,
\newblock ``Tensor decomposition for multiple-tissue gene expression experiments,''
\newblock {\em Nature genetics}, vol. 48, no. 9, pp. 1094--1100, 2016.

\bibitem{Toolbox2018}
Brett~W. Bader, Tamara~G Kolda, and Others,
\newblock ``Matlab tensor toolbox, version 2.6,''
\newblock {\em Available online at https://www.tensortoolbox.org}, 2018.

\bibitem{kossaifi2016tensorly}
Jean Kossaifi, Yannis Panagakis, Anima Anandkumar, and Maja Pantic,
\newblock ``Tensorly: Tensor learning in python,''
\newblock {\em arXiv preprint arXiv:1610.09555}, 2016.

\bibitem{he2016deep}
Kaiming He, Xiangyu Zhang, Shaoqing Ren, and Jian Sun,
\newblock ``Deep residual learning for image recognition,''
\newblock in {\em Proceedings of the IEEE conference on computer vision and pattern recognition}, 2016, pp. 770--778.

\bibitem{liu2022high}
Xiao-Yang Liu, Zeliang Zhang, Zhiyuan Wang, Han Lu, Xiaodong Wang, and Anwar Walid,
\newblock ``High-performance tensor learning primitives using gpu tensor cores,''
\newblock {\em IEEE Transactions on Computers}, 2022.

\end{thebibliography}

\end{document}